\documentclass{article}

\usepackage{amsmath,amssymb}

 \usepackage[preprint]{neurips_2023}

\usepackage[utf8]{inputenc} 
\usepackage[T1]{fontenc}    
\usepackage{hyperref}       
\usepackage{url}            
\usepackage{booktabs}       
\usepackage{amsfonts}       
\usepackage{nicefrac}       
\usepackage{microtype}      
\usepackage{xcolor}         
\usepackage{wrapfig, lipsum}
\usepackage{graphicx}
\usepackage{enumitem}
\usepackage{tabularx}
\usepackage{multirow}
\usepackage{bm}
\usepackage{booktabs}
\usepackage{tcolorbox}
\usepackage[export]{adjustbox}
\DeclareMathOperator{\Pa}{Pa}
\DeclareMathOperator{\Ch}{Ch}
\DeclareMathOperator{\De}{De}
\DeclareMathOperator{\An}{An}

\DeclareMathOperator{\EI}{\textit{EI}}
\DeclareMathOperator{\Qx}{Q_X}

\DeclareMathOperator{\Qr}{Q_R}
\DeclareMathOperator{\defi}{\overset{\textnormal{def}}{=}}
\DeclareMathOperator{\mD}{\mathbf{\textit{D}}}

\usepackage[colorinlistoftodos,size=small]{todonotes}

\newtheorem{defin}{Definition}
\newtheorem{general}{Generalization}
\newtheorem{sketch}{Sketch}
\definecolor{darkgreen}{RGB}{3,156,9}
\definecolor{royalblue}{RGB}{0,113,188}
\definecolor{aPurple}{RGB}{153,0,153}
\definecolor{aGreen}{RGB}{0,153,0}
\definecolor{aOrange}{RGB}{153,77,0}
\definecolor{aBlue}{RGB}{0,25,153}
\definecolor{aRed}{RGB}{224,0,0}

\definecolor{mapGreen}{RGB}{0,153,20}
\definecolor{mapBlue}{RGB}{0,25,153}
\definecolor{mapPurple}{RGB}{128,0,128}

\makeatletter
\newcommand\bigDiamond{\mathop{\mathpalette\bigDi@mond\relax}}
\newcommand\bigDi@mond[2]{%
  \vcenter{\hbox{\m@th
    \scalebox{\ifx#1\displaystyle 2\else1.2\fi}{$#1\Diamond$}%
  }}%
}
\newcommand\bigLozenge{\mathop{\mathpalette\bigL@zenge\relax}}
\newcommand\bigL@zenge[2]{%
  \vcenter{\hbox{\m@th
    \scalebox{\ifx#1\displaystyle 2\else1.2\fi}{$#1\blacklozenge$}%
  }}%
}
\makeatother

\title{Causal Explanations Over Time: Articulated Reasoning for Interactive Environments}

\author{%
\textbf{Sebastian Rödling}$^{*,1}$ \\
\texttt{sebastian.roedling@outlook.com}
  \And
  Matej Zečević$^{*,\dagger, {\rm 1}}$ \\
  \texttt{matej.zecevic@tu-darmstadt.de} \\
  \AND
  Devendra Singh Dhami$^{{\rm 2}}$ \\
  \texttt{d.s.dhami@tue.nl} \\
  \And
  Kristian Kersting$^{{\rm 1,3,4}}$\\
  \texttt{kersting@tu-darmstadt.de} \\
  \\
  \textsuperscript{\rm 1}Technical University of Darmstadt, Germany \\
    \textsuperscript{\rm 2}Eindhoven University of Technology, Netherlands \\
    \textsuperscript{\rm 3}Hessian Center for Artificial Intelligence (hessian.AI), Germany \\
    \textsuperscript{\rm 4}German Research Center for Artificial Intelligence (DFKI), Germany \\
  \textsuperscript{\rm *}co-first \quad  \textsuperscript{\rm $\dagger$}corresponding author\\
}

\begin{document}

\maketitle


\begin{abstract}
Structural Causal Explanations (SCEs) can be used to automatically generate explanations in natural language to questions about given data that are grounded in a (possibly learned) causal model. Unfortunately they work for small data only. In turn they are not attractive to offer reasons for events, e.g., tracking causal changes over multiple time steps, or a behavioral component that involves feedback loops through actions of an agent. To this end, we generalize SCEs to a (recursive) formulation of explanation trees to capture the temporal interactions between reasons. We show the benefits of this more general SCE algorithm on synthetic time-series data and a 2D grid game, and further compare it to the base SCE and other existing methods for causal explanations.
\end{abstract}

\section{Introduction}\label{sec:Introduction}
The progress in neural network (NN) research has led to a shift in research away from purely logic-based systems, resulting in significant changes in interpretability, capacity, and application areas. While logically formulated expressions provided interpretable results in the past, the focus now lies in incomprehensibly massive NNs. Despite notable advances in areas such as image processing \citep{ker2017deep}, text-to-image generation \citep{ramesh2021zero}, and pure text synthesis \citep{brown2020language}, it is essential to recognize the potential drawbacks associated with these models. One limitation of large models is their typical requirement for extensive training time and data. Moreover, there is an ongoing discussion in the literature about the interpretability of deep NNs, their degree of explainability, and how these factors influence users' trust in AI systems.

Potential starting points for enhancing Artificial Intelligence (AI) towards better transparency and robustness include incorporating causality (refer to the textbook by \citep{peters2017elements} for an overview of Causal AI) or continual learning (where \cite{mundt2022clevacompass} provides a comprehensive overview of this particular sub-field of AI). In this work, we aim to focus on causality, for which we provide an introduction to key ideas in section \ref{sec:Background}. In the area of Explanainable AI (XAI), several new methods have been developed for explaining predictions for different scenarios involving causality. For instance, methods such as Causal Shapley Values \citep{heskes2020causal}, causality-based counterfactuals \citep{galhotra2021explaining} or CXPlain \citep{DBLP:journals/corr/abs-1910-12336} have been developed, which can provide some notion of causality-based explanations.

Yet another approach in the AI landscape, Explanatory Interactive Learning (XIL) \citep{teso2019explanatory}, offer means to enhances user trust and model performance by interacting with users and allowing specific feedback. Furthering this, a causal variant was developed with the aim of interactively adapting a technical causal model (using the notation of Structural Causal Models, SCMs for short)) to match that of an expert's, serving as a foundation for additional systems \citep{anonymous2023causal}. In this process, a structural explanation algorithm, Structural Causal Explanation (SCE), was developed, which by design can provide truly causal explanations. This explanation algorithm generates human-readable and comprehensible responses to valid ``Why-Questions"\footnote{The term `why' is often linked to counterfactuals in causality literature. However, in the context of SCE, `why' is not directly tied to counterfactuals but is named after the framing of the questions.} within the causal model. Unfortunately the algorithm currently only supports static systems with a small amount of variables.

\begin{figure}[t!]
    \centering
    \includegraphics[width=1\textwidth]{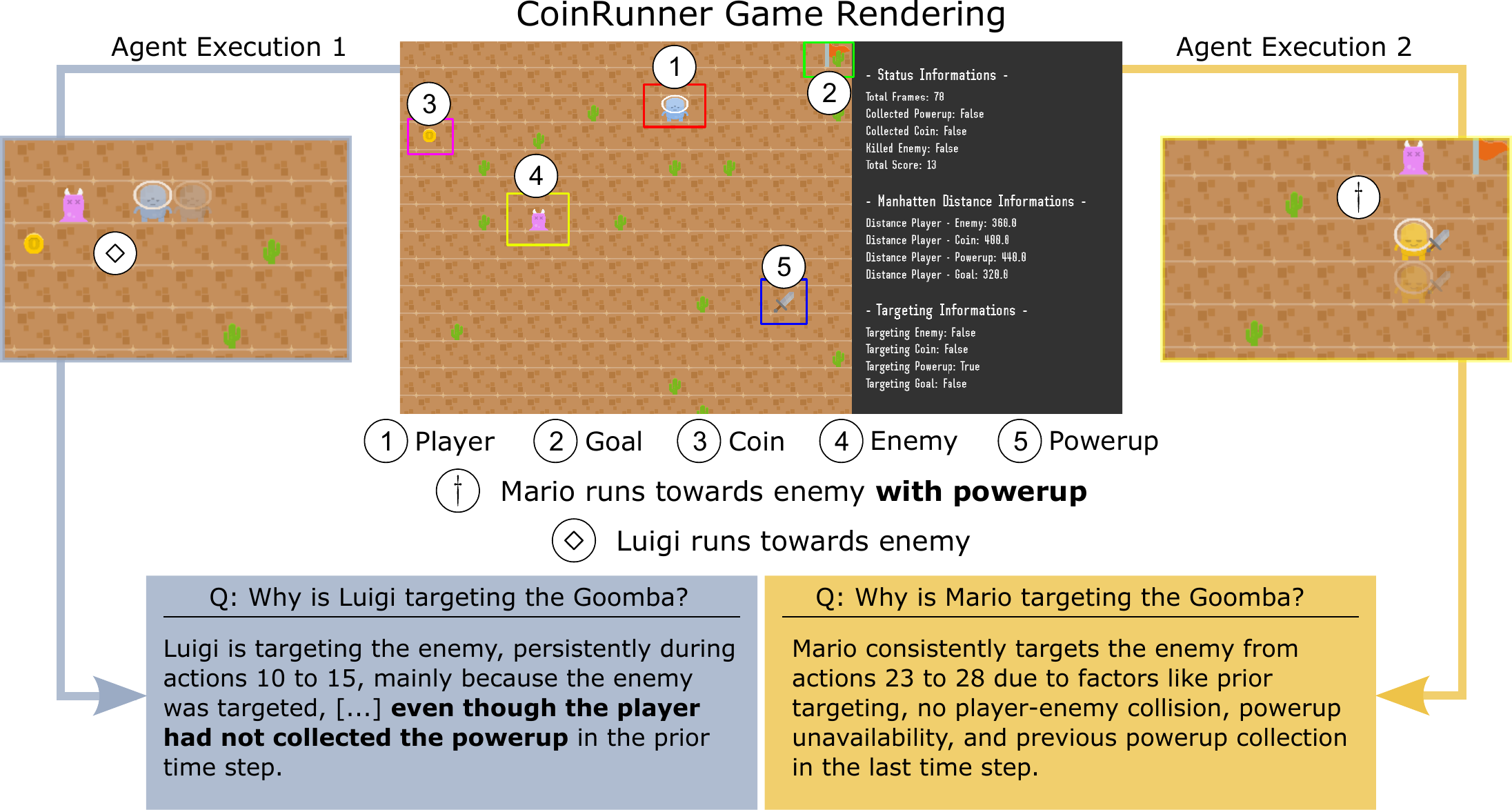}
    \caption{\textbf{``Why Was That Happening?''} Causal explanation for the behavior of two agents (referred to as Luigi and Mario) from the same player type in different game situations of the 2D CoinRunner game using the generalized T-SCE framework. (Best viewed in color)}
    \vspace{-.5cm}
    \label{fig:CoinRunner_MainFigure}
\end{figure}

While a time-independent observation is sufficient for most statistical methods, in most real-world scenarios, we will have to work with data being changed (and influenced) over time. For example, the question ``Why is the temperature at Matterhorn low?''\footnote{Matterhorn is the name of a 4500 meters tall mountain in Italy.} is being answered by SCE with ``The temperature at the Matterhorn is low because of the high altitude.'' In the so-called `Clever Hans' example, an SCM consisting of the variables `Age', `Nutrition', `Health', and `Mobility' is used to answer the question ``Why is Hans's Mobility below average?''. The answer generated by the algorithm for the in-detail discussed example is ``Hans's Mobility is bad because of his bad Health, which is mostly due to his high Age, although his Nutrition is good.'' While the first of these two examples provides a causal answer to a physical question that appears mostly time-independent, the second case addresses a question where the near or even distant time history could greatly influence the answer. It is conceivable that mobility is not only causally dependent on a person's current health but also on mobility \underline{\smash{at a previous time step}}. Similarly, current nutrition habits might depend not only on current age but also on past health. In other words, we live in a temporally measured environment and should consider this for causal explanations. It is precisely this algorithm extension that we want to address in this work. Thus, we aim to better answer the question ``Why is Hans's Mobility below average?'' and take a step further in explaining overall dynamical systems that involve not only time but even data-altering agent behavior. Figure \ref{fig:CoinRunner_MainFigure} highlights this by giving an example of two agents, whose differences in behavior become apparent through the generated explanations.

Overall, we make the following contributions: We extend the existing SCE approach to accommodate time-dependent systems, enabling better causal explanations for such systems. This extension involves introducing modifications to the underlying logic of SCE to address its shortcomings when applied to non-static data regimes and agent behaviors. Moreover, we develop a method for handling temporal dependencies within the explanation generation process. In addition to these theoretical advancements, we also evaluate our extended SCE approach on synthetic time-series data and a 2D grid game, demonstrating its effectiveness in providing more accurate and informative explanations. Finally, we compare our method with the base SCE and other existing causal explanation methods, illustrating the improvements and advantages of our proposed approach.\\\\
Our code is publically  accessible at: \href{https://anonymous.4open.science/r/Why-Was-That-Happening_Offering-Reasons-For-Events-0C4E}{Why-Was-That-Happening\_Offering-Reasons-For-Events-0C4E}

\section{Background on Causality, Time-Series and Explainable AI}\label{sec:Background}

\textbf{Structural Causal Models}: To employ causality, choosing a formalism is necessary. In this article, we use the Pearlian notation \citep{pearl2009causality}. Structural Causal Models (SCM) are defined as 4-tuples $\mathcal{M}:=(\mathbf{U},\mathbf{V},\mathcal{F},P(\mathbf{U}))$. The model consists of a set of endogenous variables $\mathbf{V}$ (all variables that have names) and a set of exogenous variables $\mathbf{U}$, which introduce stochasticity into the system through the probability function $P(\mathbf{U})$. $\mathcal{F}$ is a set of structural assignments. In typical\footnote{Markovian, where the exogenous terms are mutually independent and causal effects are identifiable.} SCM, each endogenous variable $V_i$ is determined by its causal parents $\Pa_i$ and an associated $U_i$, which can be formally expressed by the structural equation $V_i = f_i(\Pa_i, U_i)$. The mathematical definition of a causal model, along with the so-called Do-operator, offers the possibility to generate three different types of distributions, which are called associational, interventional, and counterfactual, allowing one to answer various questions like ``What does a symptom reveal about the disease?", ``What happens if I take aspirin, will my headache disappear?" or ``Was it the aspirin that relieved the headache?". These different distributions, also referred to as levels, are visualized by the well-known ladder of causality, also known as Pearl's causal hierarchy \citep{pearl2018book}.

For this work, it is particularly important that each SCM induces a causal graph $\mathcal{G}$, typically in the form of a Directed Acyclic Graph (DAG). Causal graphs can be extracted from observed data, a process known as Causal Discovery \citep{peters2017elements}. Although various approaches, such as score-based or independence-based methods, have been used to achieve this, we will not delve into their specifics since our focus is not on Causal Discovery. Notable algorithms include PC, SGS \citep{spirtes2000causation}, IC \citep{pearl2009causality}, and recent ones like NOTEARS \citep{zheng2018dags} and the continuous constrained optimization \citep{brouillard2020differentiable}. As we will see later, an approximation to an SCM, such as causal graphs, is a prerequisite for the SCE algorithm.


\textbf{Time-Series}: Time series data encompass sequences of observations captured at regular or irregular intervals. In causal models, time series data contribute to analyzing dynamic relationships between variables over time. To represent a time series, we utilize a $d$-dimensional dataset, with each entry $X_t$ comprising $d$ distinct variables $X^j_t$, expressed as the vector $X_t = (X_t^1,...X_t^d)$ for time step $t$. The time series is typically assumed to be a strictly stationary stochastic process. In time-dependent systems, the relationship structure evolves, necessitating a modification to the definition of structural assignments. Structural assignments now assume the form $X_t^j=f^j(\Pa^j_t, U_t^j)$, where $\Pa^j_t$ represents the causal parents of variable $j$ at time $t$, and $U_t^j$ corresponds to the respective exogenous influence. Analogously, we introduce the abbreviations $\Ch^j_t$, $\De^j_t$, and $\An^j_t$ to denote causal children, causal descendants, and causal ancestors, respectively. For completeness, it should be mentioned that various distributions can also be generated for time-based SCMs. The primary difference is that interventions can be carried out over a period or only at a specific time.

Similar to static Causal Discovery, algorithms for discovering causal relationships in time series data also exist. One prominent example is Granger causality, which establishes ``Granger-causal" relationships between time series based on the predictability of one series using past values of another \citep{peters2017elements}. Other methods for time series causal discovery include, i.e, VARLiNGAM, which extends the LiNGAM model to time series using vector autoregressive models (VAR) \cite{varlingam, lingam}, or the incorporation of Lasso, which employs regularization and variable selection for datasets with many variables and few examples \citep{10.1145/1281192.1281203, 10.1093/bioinformatics/btq377}.

\textbf{XAI}: Explainable Artificial Intelligence (XAI) research primarily focuses on making the decision-making process of models understandable and transparent \citep{arrieta2020explainable}. This is crucial not only to prevent erroneous decisions and ensure that individual reference points are adequately considered, but also to address ethical, legal, and societal concerns.

In the literature, a rough classification of XAI methods has been presented based on two characteristics \citep{adadi2018peeking}. First, there are methods in which (i) explanations are intrinsically embedded in the model design (e.g., decision trees) or (ii) obtained post-hoc (e.g., LIME \citep{ribeiro2016should}). Second, explanations can either be (iii) model-dependent or (iv) model-agnostic (e.g., SHAP values \citep{lundberg2017unified}). Each of these methods has its own advantages and disadvantages, making it essential to select the most appropriate method for a specific application or problem.

SHAP values, for instance, map the output to the input and generate feature importance values. LIME, on the other hand, explains individual predictions by approximating the model locally with an interpretable model. Furthermore, one can distinguish the level at which explanations are generated. Local explanations refer to individual predictions, while global explanations apply to an entire population. A third possibility, proposed by, for example, \cite{galhotra2021explaining}, is context-based explanations, which generate explanations defined by feature values for individual subpopulations.

One approach to improve explanations is to consider causality, which can provide a more in-depth understanding of the underlying mechanisms and help generate more reliable and meaningful explanations. The need for causal explanation methods is justified, among other reasons, by the assumption of feature independence in non-causal approaches. For instance, \cite{heskes2020causal} argue that non-causal methods assume feature independence when explaining predictions, which is often an invalid assumption as changing one feature can inevitably influence a correlated feature. As a result, the accuracy of these explanation methods is compromised, and they are prone to errors. In Ch. \ref{sec:evaluation}, we will delve deeper into causal XAI methods and compare them to our approach for structural explanations.

\section{Structural Causal Explanations}\label{sec:VanillaSCE}
The SCE algorithm \citep{anonymous2023causal}, designed to generate human-readable and causal explanations, can be divided into several sequential components. Most of these components have been generalized in this work and are detailed in the following section. For the sake of clarity, and to enable readers to reconstruct the original definitions, we have highlighted modifications in \textcolor{royalblue}{blue} and will provide references to these components. A critical prerequisite for the algorithm is the knowledge of a (potentially weighted) causal graph, which can be derived from data, provided by a domain expert, postulated/argued for by the data scientist or induced by an SCM.

\begin{wrapfigure}[10]{R}{0.55\textwidth}
    \centering
    \includegraphics[width=0.23\textwidth]{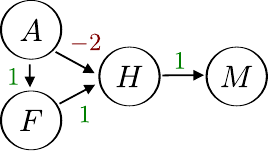}
    \caption{\textbf{Causal Graph for the Causal Hans Example.} The edge values represent the causal effect from one variable onto another. (Best viewed in color.)} \label{fig:vanillaSCESCM}
\end{wrapfigure}
If real valued weights are used, this represents some quantitative knowledge on the causal effect (which for linear SCM would be the regression coefficients). The most important aspect of the coefficients is their sign facilitating a distinction between adverse and supportive effects of a given variable onto another. Depending on the implementation of the explanation `pronunciation' procedure, cyclic graphs may also be employed. Explanations of single variables, such as self-cycles or cyclic patterns, can be filtered out during the pronounciation process.

Considering Fig. \ref{fig:vanillaSCESCM} as our causal graph, we have the variables `Age', `Food Habits', which is used synonymously with `Nutrition', `Health', and `Mobility', each denoted using their respective initials. For the SCE algorithm, initially a valid Why-Question is required. A generalized form of the Why-Question is defined in the following chapter (see Gen. \ref{def:tsscewhy}). The main difference from the original definition is the use of time steps, which is not necessary here, and the generalization from the average $\mu$ to the population statistic $\phi$. An example patient, Hans, can be represented in the form $\text{Hans}=(a_H,f_H,h_H,m_H)=(93.8,58.8,2.6,26.2)$. The question ``Why is Hans' Mobility poor?" is considered valid only if $m_H < \mu^H$ and $R(x,y):=x<y$. As noted in the original paper, any question answered in this manner is always relative to the existing population. For mobility, the terms `below average', `poor', or `immobile' are used interchangeably. For a car's fuel tank, for example, the term `low' would be used. A causal scenario (see Gen. \ref{def:tsscecs}) acts as a entry point for the Explanation Rules (see Gen. \ref{def:tssceRules}). Again, the same changes have been made. Additionally, there was an $\textnormal{\textit{ER}}3$ rule in the original version. The outcome of the Explanation Rules encapsulates the relationship between a child node and its corresponding causal parent node within an SCM. Formally, this can be expressed for each relationship of a valid question as a triplet $\textnormal{\textit{ER}}_{X \rightarrow Y} = (\textnormal{\textit{ER}}1_{X \rightarrow Y},\textnormal{\textit{ER}}2_{X \rightarrow Y}, \textnormal{\textit{ER}}3_{X \rightarrow Y})$.
The \textit{ER}1 and \textit{ER}2 rules are referred to as `Exhibition' and `Inhibition' formulations, respectively. $(\textnormal{\textit{ER}}1_{X \rightarrow Y}, \textnormal{\textit{ER}}2_{X \rightarrow Y})$ can be interpreted as ``because \{high, low\} parent" or ``although \{high, low\} parent" for the corresponding encodings $(\{-1,1\}, 0)$ and $(0, \{-1,1\})$. $\textnormal{\textit{ER}}3_{X \rightarrow Y}$ serves as an additional indicator. When a node has causal relationships with multiple variables, this rule helps identify the variable with the most significant causal influence based on scalar values. The final component is the recursion, which iterates over all variables as defined in Def. \ref{def:sci}. Consequently, given a data set and an SCM approximation, one can pose the question, ``Why is Hans' mobility poor?". The resulting explanation encodes the causal relationships underlying Hans' mobility. This explanation can be translated into human-readable language, i.e, ``Hans' mobility is poor due to his poor Health, which is primarily attributed to his high Age, despite his good Food Habits."
\begin{defin}[SCE] \label{def:sci}
Like before let $Q_X, \mathcal{M}$ be a valid why-question and some proxy SCM. Further, let $\mD\in\mathbb{R}^{n\times |\mathbf{V}|}$ denote our data set. We define a recursion 
\begin{align}\label{eq:sci-core}
\begin{split}
\mathbf{E}(Q_X, \mathcal{M}, \mD) = (&\textstyle\bigoplus_{Z\in \Pa(X)} \textnormal{\textit{ER}}(Z\rightarrow X), \textstyle\bigoplus_{Z\in \Pa(X)} \mathbf{E}(Q_Z, \mathcal{M}, \mD))
\end{split}
\end{align}
where $\bigoplus_{i=1}^n v_i = (v_1,\dots,v_n)$ denotes concatenation and \textit{ER} checks each rule \textit{ER}$i$ (Def. \ref{def:tssceRules}), and the recursion's base case is being evaluated at the roots of the causal path to $X$, that is, for some $Z{\in}\mathbf{V}$ with a path $Z \rightarrow \dots \rightarrow X$ we have
$\mathbf{E}(Q_Z, \mathcal{M}, \mD) = \emptyset.$ We call $\mathbf{E}(Q_X, \mathcal{M}, \mD)$ Structural Causal Explanation of $\mathcal{M}$.
\end{defin}

\section{Temporal - Structural Causal Explanations}


\begin{wrapfigure}[13]{R}{0.55\textwidth}
    \centering
    \includegraphics[width=0.45\textwidth]{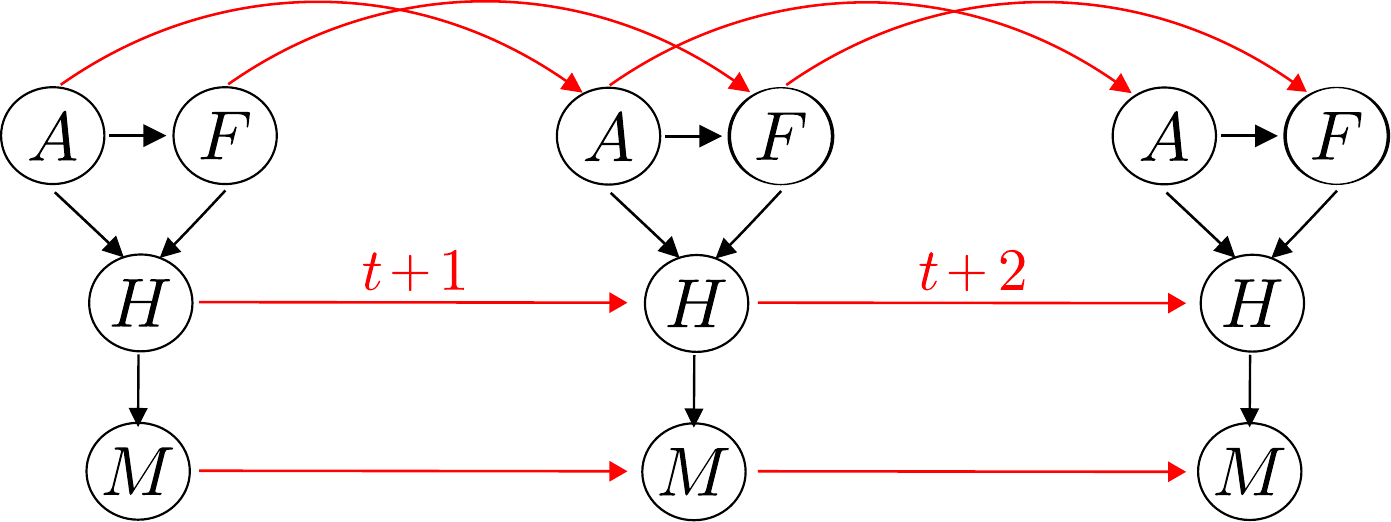}
    \caption{\textbf{Trucanated full-time graph for the Causal Hans Example.} Red edges depict a delayed effect over time. In contrast, black edges represent immediate effects. Best viewed in color.} \label{fig:T-SCE_Causal_Graph}
\end{wrapfigure}
\textbf{Incorporating Temporality}:
We expand the Causal Hans example by incorporating a time dimension into the analysis. As shown in Fig. \ref{fig:T-SCE_Causal_Graph}, our new causal time graph includes both immediate effects (black) and delayed self-effects (red). Our synthetic dataset, consisting of 10,000 records with 50 time steps each, shares a similar structure with the original dataset. Initially, the age distribution $P_A$ is uniformly distributed between 30 and 80, with each time step increasing by 1. $P_F$ is defined as $P_F = 0.5 \cdot P_A$, $P_H$ is computed using $P_H = -0.2 \cdot P_A + 0.6 \cdot P_F$, and $P_M$ is established as $P_M = 0.5 \cdot P_H$. At each time step, values are drawn from the distribution, multiplied by 0.4, and combined with the previous time step's influence, calculated with 0.6. To maintain consistency, each distribution's mean is determined, and noise is introduced during new distributions' evaluation through $+\mathcal{N}(0,\mu^P \cdot 0.03)$.

\begin{general}[Why Question] \label{def:tsscewhy}
For individual $i$ and their instance $x_t^i \in \text{Val}(X_t)$ of any variable $X_t\in\mathbf{V}$ in SCM $\mathcal{M}$ at time point $t$, let $\textcolor{royalblue}{\phi^{X_t}}$ be a population statistic (e.g., mean, 10\% percentile). A why-question $Q_{X_t^i} \defi R(x_t^i, \textcolor{royalblue}{\phi^{X_t}})$, with $R\in \{<, >\}$, is true if $Q_{X_t^i}$ holds.
\end{general}
\begin{general}[Causal Scenario (CS)] \label{def:tsscecs}
$C_{XY}\defi(\alpha_{X\rightarrow Y}, x_t, y_k, \textcolor{royalblue}{\phi^{X_t}}, \textcolor{royalblue}{\phi^{Y_k}})$ is called a CS.
\end{general}
\begin{general}[Explanation Rules] \label{def:tssceRules}
Let $C_{XY}$ denote a causal scenario. Given a sign function $s(x) \in \{-1,1\}$, a binary ordering relation $R_i\in \{<,>\}$, we define FOL-based rule functions ER$i(\cdot)\in\{-1,0,1\}$, indicating how the causal relation $X{\rightarrow} Y$ satisfies each rule. \textcolor{royalblue}{ER$i(\cdot)$ evaluates both Fundamental Rules and situational Complementary Rules, which can be added.} \textcolor{royalblue}{Here, $t$ and $k$ are time steps, where $t \ge k$}

\begin{enumerate}[label={(ER\arabic*{})},start=1]
\item
$
\begin{aligned}[t]
\textnormal{If } R_1 \neq R_2 \textnormal{, then: }
&((s(\alpha_{X {\rightarrow} Y}) < 0)\land \delta_1) \vee ((s(\alpha_{X {\rightarrow} Y})> 0) \land \delta_2 )
\end{aligned}
$
\item
$
\begin{aligned}[t]
\textnormal{If } R_1 \neq R_2 \textnormal{, then: }
&((s(\alpha_{X {\rightarrow} Y}) > 0)\land \delta_1)\vee ((s(\alpha_{X {\rightarrow} Y}) < 0) \land \delta_2)
\end{aligned}
$
\end{enumerate}
\begin{center} with
$\delta_1 \defi R_2(x,\textcolor{royalblue}{\phi^{X_t}}) \land R_1(y,\textcolor{royalblue}{\phi^{Y_k}}))$ and $\delta_2 \defi R_2(x,\textcolor{royalblue}{\phi^{X_t}}) \land R_2(y,\textcolor{royalblue}{\phi^{Y_k}}))$
\end{center}
\end{general}
\begin{defin}[Explanation Tree] \label{def:explanation_tree}
An Explanation Tree is a directed acyclic graph (DAG) $\mathcal{T}$. The root node $r \in \mathcal{N}$ represents the variable of interest (valid Why-Question). Each edge $(u, v) \in \mathcal{E}$ is directed from parent node $u \in \mathcal{N}$ to child node $v \in \mathcal{N}$, such that in the retrospective case, the children of a node in the tree represent the causally explanatory variables of their parent node. In the anticipative case, the relationship is reversed, with the child nodes being explained by their parent node in the tree.
\end{defin}
As before, we focus on the patient Hans. His data can now be represented at each time step $t$ as a tuple $X_t = (X^A_t, X^N_t, X^H_t, X^M_t)$. Valid questions are defined according to Gen. \ref{def:tsscewhy}, and the causal scenario (Gen. \ref{def:tsscecs}) aligns with the original concept. The Explanation Rules (Gen. \ref{def:tssceRules}) have been adapted to accommodate time series data while still preserving the underlying idea that these rules encode causal relationships. Notably, variables $X$ and $Y$ can now originate from two distinct time steps. While the original algorithm used a linked list as its fundamental data structure and produced nested sentences for explanations, we have chosen to extend it with a tree data structure (Def. \ref{def:explanation_tree}). As the number of explanatory variables increases over time and for larger SCMs, nested explanations inevitably lead to confusion. Moreover, our switch to the new data structure implies that we aim to provide one-sentence explanations for each variable to be explained. As we will see later, this also has advantages for summarizing causal relationships over time or identifying changing relationships. To construct the Explanation Tree, the SCE recursion must be adjusted accordingly (see Def. \ref{def:tssce}).
\begin{defin}[Temporal SCE] \label{def:tssce}
Let $\Qr$ be the root node of a valid why-question and $\mathfrak{M}$ a set of proxy SCMs for distinct contexts. Moreover, let $\mD\in\mathbb{R}^{n\times |\mathbf{V}|}$ represent our dataset and $K \in \mathbb{N}$ denote the maximum recursion depth, with the starting depth being $j = 0$. In the first iteration, $\Qx = \Qr$. We define a recursion as follows:

\noindent\begin{equation}
\mathbf{E}(\Qr, \Qx, \mathfrak{M}, \mD, j) = \Big(\bigoplus_{Z\in \Phi_1} \EI(Z,X_t), {\bigDiamond_{Z\in\Phi_1\setminus{\Phi_2 }}}\begin{cases}
\mathbf{E}(\Qx, Z, \mathbf{\mathfrak{M}}, \mD, j + 1), & \textnormal{if } j < K \\
\emptyset, & \textnormal{else}
\end{cases}\Big)
\end{equation}
$\bigoplus$ attaches new nodes to the given node depending on the used case. Explanation Indicators, $\EI(\cdot)$, resolves all \textit{ER}$i$ and stores the time step $t$, variable names and context of the nodes. The iterator $\textstyle\bigDiamond$ maintains the recursion over included nodes. The $\Phi_1$ further select the appropriate approx. SCM from $\mathfrak{M}$ based on the dataset, time step, and variable (context's). The set $\Phi_2$ prevents from reattaching duplications by checking for existing time $t$ and variable name combinations. 
\end{defin}


\begin{defin}[Retrospective]\label{def:retrospective}
If $\Phi_1\defi (\Pa^X_t \cup \Pa^X_{\textnormal{t},\theta} \cup \Pa^X_{\textnormal{t},n})$ and $\Phi_2\defi\De^{\Qr}$ where $\Pa^X_{t,\theta} \defi \{Z \in \Pa^X_t \mid |\alpha_{Z{\rightarrow} X}| > \theta\}$ and $\Pa^X_{t,n} \defi \{\textnormal{top}n(Z \in \Pa^X_t, |\alpha_{Z{\rightarrow} X}|)$ then we call $\mathbf{E}$ retrospective.
\end{defin}

\begin{defin}[Anticipative]\label{def:anticipative}
If $\Phi_1\defi (\Ch^X_t \cup \Ch^X_{\textnormal{t},\theta} \cup \Ch^X_{\textnormal{t},n})$ and $\Phi_2\defi\An^{\Qr}$ where $\Ch^X_{t,\theta} \defi \{Z \in \Ch^X_t \mid |\alpha_{X{\rightarrow} Z}| > \theta\}$ and $\Ch^X_{t,n} \defi \{\textnormal{top}n(Z \in \Ch^X_t, |\alpha_{X{\rightarrow} Z}|)$ then we call $\mathbf{E}$ anticipative.
\end{defin}

\begin{wrapfigure}[14]{R}{0.55\textwidth}
    \centering
    \includegraphics[width=0.5\textwidth]{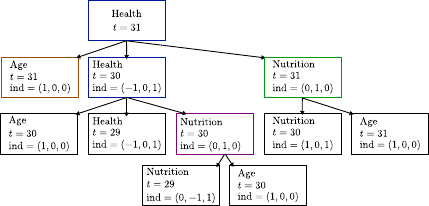}
    \caption{\textbf{Example of Explanation Tree.} Sub-tree regarding the question, ``Why is Hans' Health below average?''. Sequences are color-coded.} \label{fig:explanationTree}
\end{wrapfigure}
Def. \ref{def:tssce} is used to construct an Explanation Tree $\mathcal{T}$ for a valid question. Additionally, a distinction is made between the retrospective and anticipatory cases. Given time series and a causal graph, we can provide explanations for the emergence of our variable of interest at a specific point (retrospective reasons, see Def. \ref{def:retrospective}). Additionally we can include causal effects as explanatory factors, e.g., when we aim to explain behavior (anticipative effects, see Def. \ref{def:anticipative}). Furthermore, these case differentiations are designed to narrow down the possible number of explanatory variables and select the appropriate SCM among different options for a given point in time. For our time series Causal Hans example, we currently focus on one context (SCM) and retrospective explanation.  The data structure now allows for further indicators or manipulations to be performed directly on the tree. A sequence indicator, which helps to mark sequences of consistent causal relationships, is of crucial importance for summarization and identification of changing relationships. Masking variables (e.g., Age, as the causal explanation is generally given) or trimming down the tree to individual path explanations are also possible. To implement the retrospective sequence indicator, we need to define a function $f(\mathcal{T})$ that iterates separately over all endogenous variables of the SCMs in $\mathcal{T}$ in descending temporal order. If direct child nodes have the same Explanation Rule indicators and remain stable over time, a unique ID is assigned to the sequence. Leaves should be excluded, since they are just explanatory and not to be explained.


In the following, we provide an example illustrating human-readable causal explanations in response to the question ``Why is Hans' Health below average?'' with the recursion depth set to 2. The varying explanations for Hans' nutrition and the perfectly summarizable explanation for his health are highlighted in color. It should be noted once more that effects from the past are always positive, and as in the original causal Hans graph, the effect of age on nutrition is also positive, whereas age has a negative impact on health. The corresponding Explanation Tree is shown in Fig. \ref{fig:explanationTree}. A more detailed example can be found in the appendix.

\begin{tcolorbox}[sharp corners, boxrule=1pt, colback=white, colframe=black, boxsep=2pt, left=2pt, right=2pt, top=2pt, bottom=2pt]
Hans' \textcolor{mapBlue}{Health has consistently been below average over the last two years}, mostly because his low Health persistently one year prior and because of his high Age in the referenced year, despite his high Nutrition in the referenced year.  \vspace{1mm}\\
His \textcolor{mapGreen}{Nutrition was above average this year}, due to his high Nutrition one year before and his high Age in the same year. \vspace{1mm}\\
\textcolor{mapPurple}{Last year, his Nutrition was above average}, because of his high Age in the same year, despite his low Nutrition one year before.
\end{tcolorbox}


\textbf{Dynamic Time Series}: It is entirely conceivable that our time-series Causal Hans example does not accurately reflect reality and that we can depict this more precisely. Theoretically, it could be that the relationship between Age and Nutrition during adolescence ($\textnormal{Age} \leq 25$) is primarily negative, which could be due to the consumption of alcohol or junk food, and subsequently ($\textnormal{Age} > 25$) turns positive again. In this scenario, two SCMs could be deemed valid based on a specific variable, thereby allowing for a seamless transition between the SCMs and the variables used within them. The intuition behind Gens. \ref{def:tsscewhy}, \ref{def:tsscecs} \ref{def:tssceRules}, and Def. \ref{def:tssce} would still be applicable in this case. It is crucial to employ the accurate causal graph and the appropriate data, as outlined in Def. \ref{def:tssce} (see $\phi_1$ and $\mathfrak{M}$). The Explanation Tree and the resulting explanations would continue to operate effectively, allowing for the Age variable to be accurately summarized in each instance. However, the explanation may evolve over time for some examples due to varying causal relationships.

We now turn our attention to agent behavior in a 2D grid-based game called CoinRunner (see Fig. \ref{fig:CoinRunner_MainFigure}), inspired by CoinRun \citep{pmlr-v97-cobbe19a}. The game comprises various sprites, including Goldcoin, Powerup, Enemy (referred to as Goomba), Player (Mario or Luigi), and Goal. The player starts with 20 points and loses one point per second until they reach the Goal. Collecting Goldcoins grants 5 bonus points (BP), while colliding with the Enemy yields 9 BP if the Powerup has been collected beforehand, and results in a game over with a score of -20 if no Powerup was obtained. The initialization and positioning of the sprites are independently randomized. Each game rollout $R$ is represented by a sequence of frames $f$, each described by mostly binary variables.

The main difference from the Causal Hans example is that the focus is now on explaining an agent's behavior within a rollout rather than examining population statistics. Interestingly, it is intuitive that explaining behavior involves not only retrospective considerations but also anticipatory reasons and consequences of specific actions. To this end, Def. \ref{def:tssce} has an anticipatory case added (Def. \ref{def:anticipative}), which can be used to encode effects in the future within the current context when the future is still uncertain, or across contexts when the future has already occurred (e.g., ``Colliding with the Enemy positively impacts defeating the Enemy at the time step."). The abstraction of the T-SCE defs. to the binary case is rather direct and does not require much elaboration. Further details are provided in the appendix.

For this game, we can define several deterministic agent behaviors, such as Coincollector, Killer, or Optimal. For the purpose of demonstration, we will highlight the Killer agent behavior. This behavior primarily focuses on collecting the Powerup and colliding with the Enemy when both Powerup and Enemy are present, before proceeding to the Goal. Similar to the dynamic Causal Hans example, various stationary processes can be identified from this. These primarily depend on the game's state (variables of a frame). For our Killer agent, processes can be most easily described based on the existence of sprites. We call $C$ a context, which brings us to the currently valid SCM and thus to a stationary subprocess. Specifically, for the Killer agent, we have defined the following three subprocesses: (i) $C_{K,1} = \textnormal{powerup and opponent exist'}$, (ii) $C_{K,2} = \textnormal{powerup does not exist and opponent exists'}$, and (iii) $C_{K,3}=\textnormal{`neither exists'}$. For this purpose, we have implemented an imperfect Killer agent and recorded 500 rollouts frame by frame. By imperfect, we mean that, with very low probability, it can also exhibit other behavior. Together with a bit more noise, we then used Lasso, VARLiNGAM \citep{varlingam}, and Granger on conditioned frame sections, depending on our contexts $C$, to generate graphs that we want to assume as causal for the moment. The learned graphs, a description of the methods used, and further T-SCE parameters are not essential for the main part and the demonstration of the application of this work and have also been moved to the appendix.

As a result, we can already causally explain questions like ``Why did Mario jump on the Goomba?", ``Why is Mario targeting the Goomba?" or ``Why does Mario run into the Goal?", if the question is valid in the specific time step and, as in our case, at runtime. For this purpose, the current frame is used to identify the currently valid SCM, and the rest of the recorded dataset serves as the  explanation basis. Fig. \ref{fig:CoinRunner_MainFigure} shows a scenario in which Mario is running towards the opponent with the retrospective explanation on  Fig. \ref{fig:CoinRunner_MainFigure} (right). The corresponding \emph{anticipatory} explanation is: 
\begin{tcolorbox}[sharp corners, boxrule=1pt, colback=white, colframe=black, boxsep=2pt, left=2pt, right=2pt, top=2pt, bottom=2pt]
Targeting the enemy has a positive effect on targeting the enemy, the existence of the enemy, colliding with the enemy and a negative effect on the score, targeting the goal and killing the enemy in the next time step.
\end{tcolorbox}

\section{Contexting T-SCE with Existing Causal XAI Paradigms}\label{sec:evaluation}
\begin{figure*}[t]
\centering
    \includegraphics[width=1\textwidth]{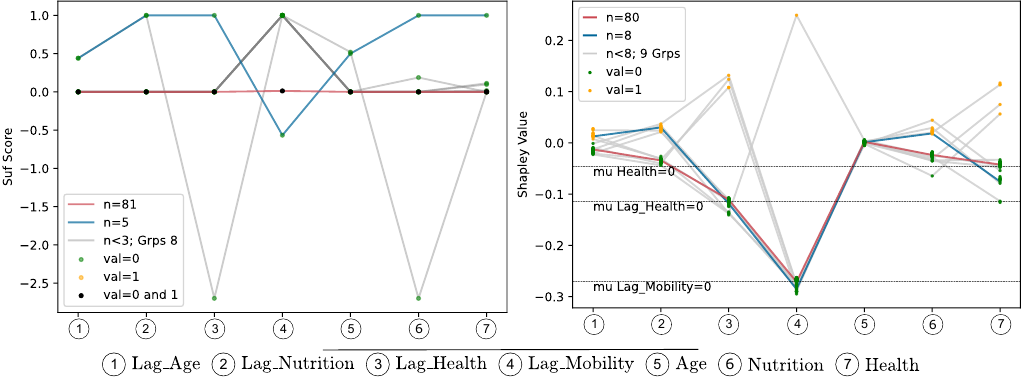}
    \caption{\textbf{SotA methods are not designed to offer structural explanations.} Suf. Score and Shapley Values for 100 random patients with below avg. mobility prediction. Same Score/Value sequences are grouped together (group size n). Small groups are again combined.}
    \vspace{-.25cm}\label{fig:contextJoint}
\end{figure*}

In this section, we aim to compare T-SCE with other state-of-the-art (SotA) causal explanation methods and, of course, base SCE. Since no method is available that generates both \textit{structural} and causal explanations over time, we must limit ourselves to a very asymmetric comparison. Furthermore, each method is based on different assumptions and is designed for different purposes. Our method relies on explaining through the structural effect direction (and possibly strength), rather than the combination of feature value and structural influence of individual attributes. For evaluation purposes, we are interested in the question ``Why is Hans' Mobility below average?".

For the comparison, we have adapted the synthetic Causal Hans time series dataset so that the causality-based counterfactuals \cite{galhotra2021explaining} and Causal Shapley Values \cite{heskes2020causal} can be used similarly.  The new dataset is designed as a classification problem and contains the categorization `below average' (0) or `above average' (1) for all variables across all patients and time steps. In line with our causal graph \ref{fig:T-SCE_Causal_Graph}, we have added delayed variables and immediate variables as `lagged' (prefix `Lag\_') and `non-lagged' respectively in each sample.

The causality-based counterfactual explanation method LEWIS enables the calculation of Sufficiency- (Suf), Necessity- (Ne), and NeSuf-Score(s) to provide explanations at various levels (local, contextual, global) for a binary classification (e.g., credit approval). Depending on the feedback (0 or 1), a different value (Suf or Ne) is important for reversing the feedback with a certain probability (calculated score).
In terms of our classification problem, this signifies that a prediction of 0 (below average) is depicted by the Sufficiency Score of each variable, illustrating how each could potentially contribute to a shift in mobility. As we are interested in the extent to which the current value is used for explanation, we choose the alternative category as the starting point and calculate a transition to the current category. Contrary to our method, we obtained diverse explanatory variables (and values) for the local explanations here. Fig. \ref{fig:contextJoint} (left side) presents the Suf Scores for 100 patients with prediction below average mobility. For our dataset, Suf Scores in the range of 0 were often calculated across all variables for several patients. In isolated cases, variables received differing Suf Scores, with the most variation in 'Lag\_Mobility'.
Ultimately, the method is not designed to distinguish direct from indirect effects or to generate consistent explanation variables across different patients. Moreover, the evaluation of whether a category of a feature has a positive or negative effect is only possible through trial and error, insofar as there is no prior knowledge about the categories.

To compare with the Causal Shapley Values, we used a similar setup. The Causal Shapley Values differ significantly from other Shapley Values (\citep{aas2020explaining, janzing2019feature, frye2021asymmetric}
), as they are able to quantify the influence of direct and indirect variables differently. The inclusion of a partial order of the real causal structure underlying the data limits the permutation possibilities for calculation. The authors have introduced both symmetric and asymmetric Causal Shapley Values. Since asymmetric values tend to place more weight on the root element (`Lag\_Age' in our case) in chain graphs, and we are more interested in direct and indirect effects, we used symmetric values for our experiment, which include both influences in the calculations and distribute the effect. For testing purposes, we provided a partial order corresponding to the temporal sequence of our causal graph ([\{`Lag\_Age', `Lag\_Nutrition', `Lag\_Health', Lag\_Mobility'\},\{Age, Nutrition\},\{Health\}]) and plotted the Causal Shapley Values for 100 patients with below average mobility in Fig. \ref{fig:contextJoint}. It is evident that 'Lag\_Mobility' exerts a significant influence. The direct effect of 'Health' is likely diminished in this partial order due to its involvement in many indirect effects. Notably, within this order, 'Lag\_Health' assumes greater importance than 'Health' in the predictions. Whether a (truly direct) variable generally has a positive or negative effect on the target variable in this case can be straightforwardly determined by comparing it with the complementary group 'above average.' For our case, this roughly corresponds to a reflection along the x-axis and an inversion of colors. The resulting values can be used to explain individual predictions, but it requires considerable additional effort in the analysis to focus exclusively on direct effects. Causal Shapley Values do incorporate causal structure but are also not directly designed to give structural causal explanations. 


T-SCE is an extension of the original SCE algorithm. In our work, we have generalized the definitions of SCE and made them suitable for explaining time series. T-SCE has gained the ability to summarize explanations for stationary time series models and to limit the number of explanatory variables in a situation-specific manner. During the development of T-SCE, care was taken to ensure that the definitions remain backward compatible, so that both temporal and more static explanation encodings can be generated. T-SCE inherits some weaknesses of the SCE algorithm in terms of applicability to linear models, as well as the strengths of intuitive and simple structural explanations, which are suitable, e.g., for the causal XIL paradigm. T-SCE would answer our question of interest, i.e., primarily like followed (apart from the individual explanations of the explanatory variables):

\begin{tcolorbox}[sharp corners, boxrule=1pt, colback=white, colframe=black, boxsep=2pt, left=2pt, right=2pt, top=2pt, bottom=2pt]
Hans' Mobility has been conistently below average for the past three years \textcolor{royalblue}{due to his poor Mobility one year prior} and \textcolor{mapPurple}{ his weak Health in the current year}.
\end{tcolorbox}

\section{Concluding Remarks}


We have improved upon the original SCE algorithm by expanding its capability from providing causal and structural explanations to include temporal explanations as well. This progression also allows for a nuanced differentiation between anticipatory and retrospective aspects. This refined T-SCE algorithm now enables a flexible limitation of explanatory variables, proving to be especially beneficial in situations involving learned causal graphs with numerous minimally influential variables. For the first time, T-SCE has been illustrated to provide causal explanations to agent behavior in a 2D game, corroborating its wide applicability. To enhance user trust, a domain expert can either provide a causal graph or use a learned one to then interact with the provided explanations, incrementally refining the given model before its application \citep{anonymous2023causal}. These incrementally improved SCMs can potentially be integrated into other systems, for example, as a regularization term, further demonstrating the adaptability and versatility of T-SCE. While T-SCE inherits the linearity assumption of SCE, this comes with the benefit of providing uniquely structured explanations.

We have explored the state-of-the-art methods for causal explanations, namely LEWIS and Causal Shapley Values, each demonstrating unique merits under varying circumstances. LEWIS is engineered to interact with feedback from classification models, while Causal Shapley Values adeptly contrast the significance of individual features, computing their influence on the target variable given partial causal knowledge. Standing apart, T-SCE presents a unique capacity for delivering structural explanations, applicable in both static and temporal contexts, thus proving itself to be a useful tool in the realm of causal reasoning.


\bibliographystyle{alpha}
\bibliography{neurips_2023}

\clearpage
\appendix

\section*{Supplementary Material to\\ ``Why Was That Happening?'' Offering Reasons For Events}

Our code is publicly accessible through anonymous.4open.science at the following link: \href{https://anonymous.4open.science/r/Why-Was-That-Happening_Offering-Reasons-For-Events-0C4E}{Why-Was-That-Happening\_Offering-Reasons-For-Events-0C4E}

The code was executed on a Windows 11 Pro computer with an AMD Ryzen 9 3900X (12-Core, 3800 MHz) processor and 32GB of RAM. The computer is equipped with an NVIDIA GeForce GTX 1070 graphics card.

\section{Towards Temporal SCE - Causal Hans}
The main body of the work primarily discusses the theoretical background of the T-SCE algorithm. Here, we aim to present a more practical perspective, illustrating the entire process of causal explanation generation. 
\begin{figure*}[h]
    \centering
    \begin{adjustbox}{max width=1.7\linewidth,center}
    \includegraphics[scale=1.2]{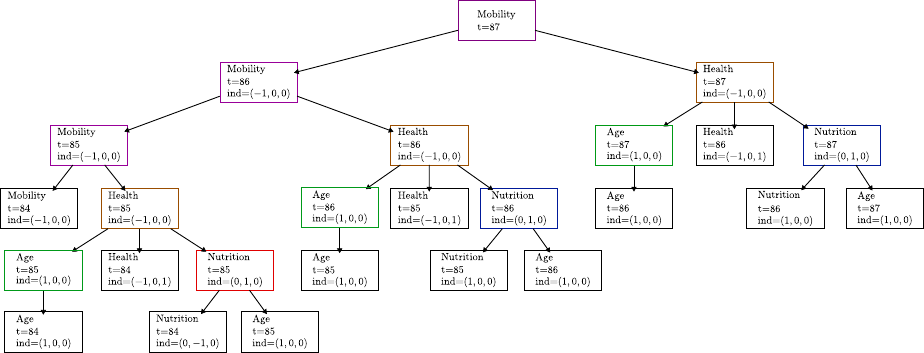}
    \end{adjustbox}
    \caption{\textbf{Explanation Tree for the Causal Hans Example with changing explanations over time.} Visualized are the individual nodes together with their indicators (ind) for the Fundamental Rules, Complementary Rule and the time (t). Color-coded are sequences tags indicating that the explanatory tree child nodes have continuously the same indicators for the parent node. Note `Nutrition' changes color over time. Leaf nodes are not further individually explained.} \label{fig:tssce_explanation_tree_changing_relations}
\end{figure*}

We continuously refer back to the Time Series Causal Hans dataset and causal graph (Fig. \ref{fig:T-SCE_Causal_Graph}) and evaluate the well-known retrospective question ``Why is Hans' Mobility below average?" (For this example, we assume the question is valid for $t=87$). In this case, we have limited the analysis to a recursion depth of 3 and included the complementary rule $ER3$ (Mostly-Rule). The question also reveals that our Why-Question (Gen. \ref{def:tsscewhy}) refers to the $\phi$ corresponding to the 'average.' Furthermore, since there is a manageable number of variables, we use all causal parents. Fig. \ref{fig:tssce_explanation_tree_changing_relations} shows the corresponding Explanation Tree for our scenario, which we will now go through step by step. We can pass the valid question (Root Node) to our T-SCE recursion (Def. \ref{def:tssce}). The direct causal parents of the Mobility variable from $t=87$ are added as new child nodes in the explanation tree. The evaluation of the $\EI(\cdot)$ (Explanation Rules Gen. \ref{def:tssceRules}) and the time step $t$ of the corresponding variable are recorded. The structural causal parents are ``Mobility" from time step $t=86$ and Health from $t=87$. Each tree child node also contains, in our case, the triplet $\textnormal{\textit{ER}}_{X \rightarrow Y} = (\textnormal{\textit{ER}}1_{X \rightarrow Y},\textnormal{\textit{ER}}2_{X \rightarrow Y}, \textnormal{\textit{ER}}3_{X \rightarrow Y})$, which encodes the relationship to the parent node. The process is continued recursively across the tree children until the last recursion encounters the empty set. Post-hoc, we can search for consistent relation encoded sequences using a sequence indicator function. Since causal time series models assume a stationary process, which means that the data generation process (in our case, given by the structural equations of the SCMs) do not change, we do not have to think further about different time jumps of causal parents and or changing causal parents within a context. Instead, we are only interested in the encoded causal relationships over time and whether they change. In Fig. \ref{fig:tssce_explanation_tree_changing_relations}, sequences are indicated by color markings. For example, Mobility is continuously explained by Mobility from the previous time step and Health from the same time step, with the same relationship encoding over time. Nutrition, on the other hand, contains a changing explanation, which is not due to the structural equations but to the noise described in the system. Consequently, we can decode the relationships and information encoded in the Explanation Tree into the following human-readable language, for example:
\begin{tcolorbox}[sharp corners, boxrule=1pt, colback=white, colframe=black, boxsep=2pt, left=2pt, right=2pt, top=2pt, bottom=2pt]
Hans' \textcolor{aPurple}{Mobility} has been \textcolor{aPurple}{consistently below average over the past three years} mostly due to his low Mobility continuously one year before and his low Health in the current year.
\vspace{1.5mm} \\
His \textcolor{aOrange}{Health} has been \textcolor{aOrange}{consistently below average over the past three years} due to his high Age in the current year and mostly because his low Health continuously one year before, although his Nutrition was good in the current year.
\vspace{1.5mm} \\
His \textcolor{aBlue}{Nutrition} has been \textcolor{aBlue}{consistently above average for the past two years} \textcolor{black}{due to his good Nutrition in the previous year} and his high Age in the current year.
\vspace{1.5mm} \\
\textcolor{aRed}{Two years ago, his Nutrition} was above average because of his high Age in the same year, \textcolor{black}{despite his low Nutrition the year before}.
\vspace{1.5mm} \\
His \textcolor{aGreen}{Age} has been \textcolor{aGreen}{consistently above average for the past three years} \textcolor{black}{due to his high Age in the previous year}.
\end{tcolorbox}
Depending on the situation, it may be that not the entire tree should be used for the explanation, and changes other than limiting the width or depth of the tree should be made. We present three informal operation options that (i) trim the tree to a path of a certain width to understand the structural influence of another variable on the target variable, (ii) mask intermediate nodes for linear graphs if they do not require an explanation, and (iii) manipulate short sequence jumps within an otherwise consistent long sequence for particularly noisy data using the sequence indicator (e.g., for explaining only marco changes).
\begin{sketch}[Path-based Explanations with Channel Width] \label{def:path_based_explanations_channel_width}
Given an Explanation Tree $\mathcal{T} = (\mathcal{N}, \mathcal{E})$ with root node $r \in \mathcal{N}$ and a unique directed path $P$ from the root to a non-leaf node, a Path-based Explanation with Channel Width $w$ is a subgraph $\mathcal{T'} = (\mathcal{N}', \mathcal{E}')$ of $\mathcal{T}$, where $\mathcal{N}'$ is the set of nodes in the path $P$ and their children up to a depth of $w$, and $\mathcal{E}' \subseteq \mathcal{E}$ is the set of directed edges connecting nodes in $\mathcal{N}'$. The Path-based Explanation with Channel Width $w$ focuses only on the variables in the unique path $P$ and their children up to a depth of $w$ for providing explanations.
\end{sketch}

\begin{sketch}[Masking Operation] \label{def:masking_operation}
Given an Explanation Tree $\mathcal{T} = (\mathcal{N}, \mathcal{E})$ and a subset of nodes to be masked $\mathcal{M} \subseteq \mathcal{N}$, the Masking Operation produces a new Explanation Tree $\mathcal{T}' = (\mathcal{N}', \mathcal{E}')$ as follows:
\begin{align*}
\mathcal{N}' &= \mathcal{N} \setminus \mathcal{M}, \\
\mathcal{E}' &= { e_{ij} \in \mathcal{E} \mid n_i \notin \mathcal{M} \wedge n_j \notin \mathcal{M} } \\
&\quad \cup { e_{ik} \mid n_i, n_k \notin \mathcal{M} \wedge \exists P_{ik} \subseteq \mathcal{N} \setminus \mathcal{M} },
\end{align*}
where $P_{ik}$ is a minimal path from $n_i$ to $n_k$ in $\mathcal{T}$ that does not contain any other nodes in $\mathcal{M}$. The weight $w_{ik}$ of each new edge $e_{ik}$ is set to the product of the weights of the original edges in the path $P_{ik}$. For each new edge $e_{ik}$, evaluate it against the given Fundamental/Complementary Rules to update the corresponding child node $n_k$. The Masking Operation only applies to intermediate nodes and not to leaf nodes, as leaf nodes do not have any further causal relationships to be preserved.
\end{sketch}
\begin{sketch}[Leave-N-Out Method] \label{def:leave_n_out_method}
Given an Explanation Tree $\mathcal{T} = (\mathcal{N}, \mathcal{E})$, a set of interrupted sequences $I \subseteq \mathcal{N}$, and a function $f$ that updates the sequence ID and meta information for the Fundamental/Complementary Rules, the Leave-N-Out Method modifies the Explanation Tree to merge interrupted sequences and update the related nodes' meta information. The updated Explanation Tree $\mathcal{T}' = (\mathcal{N}', \mathcal{E}')$ is defined as follows:
\begin{enumerate}
\item For each node $n_i \in I$ with an interrupted sequence:
\begin{enumerate}
\item Identify the node $n_j \in \mathcal{N}$ that precedes $n_i$ in the original sequence, and the node $n_k \in \mathcal{N}$ that follows $n_i$ in the resumed sequence.
\item Update the sequence ID of $n_i$ to match the sequence ID of $n_j$.
\item Optional: Update the meta information of $n_k$ using the function $f$, ensuring that the updated node $n_k$ includes the updated Fundamental/Complementary Rules information, so that $n_i$ gets explained according to the original sequence.
\end{enumerate}
\item Define the updated Explanation Tree $\mathcal{T}' = (\mathcal{N}', \mathcal{E}')$, where $\mathcal{N}' = \mathcal{N}$ and $\mathcal{E}' = \mathcal{E}$, with the nodes in $\mathcal{N}'$ containing the updated object-oriented class instances and meta information.
\end{enumerate}
\end{sketch}

\section{Towards Temporal SCE - CoinRunner}
The CoinRunner example has contributed only a relatively small proportion to the main body of this work. We intend to use this appendix to provide more detailed information about the game, the learned causal graphs, and additional explanatory examples, as well as to discuss the minor definition modifications required for the implementation of causal explanations.

\begin{figure*}[h]
\centering
    \includegraphics[width=0.85\textwidth]{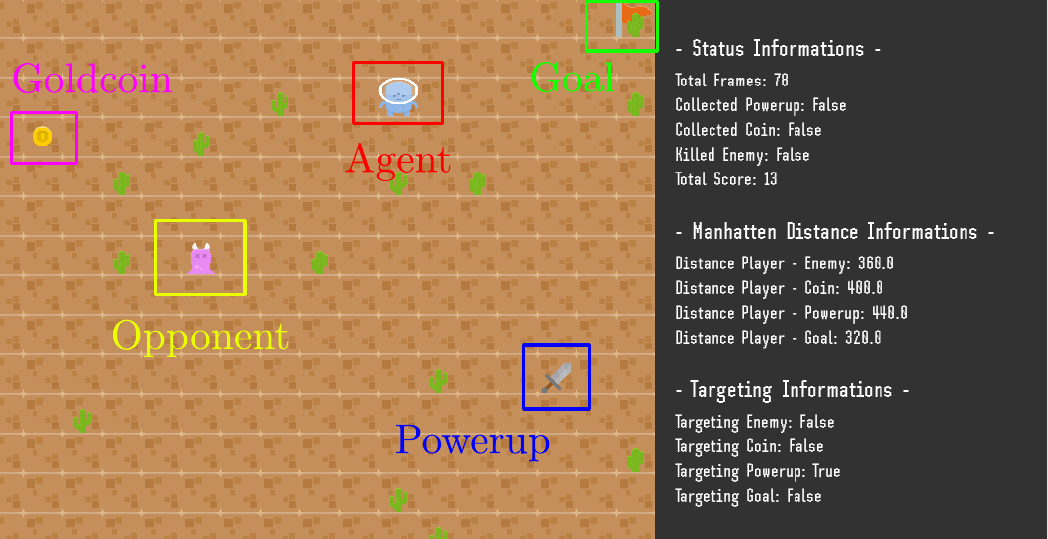}
    \caption{\textbf{Basic CoinRunner game board including information board on the right side.} Agent, gold coin and powerup have been initialized and are placed at random positions in the environment. The target sprite is placed at one of the four corners. Best viewed in colors.} \label{fig:basicCoinrunner}
\end{figure*}

Fig. \ref{fig:basicCoinrunner} displays the scope of the implemented game as described in the main body of this work. On the left side is the playing field with all possible individual sprites. The agent is currently (still) represented as a blue avatar. The yellow version of the avatar indicates that the power-up has been collected. On the right side is an information board, which is not of further interest to us in the context of this work.

To provide valid questions and meaningful encoding of causal relationships, we will first adapt the definitions of the T-SCE for our use case. Key differences include that we are no longer interested in population statistics, but rather in the agent's behavior during a rollout. Anticipatory explanations were already discussed in the main body, and a definition was provided. \textit{At runtime} of a rollout as in our case this is limited to a pronounciation like 'has a positive' or 'has a negative effect', affecting the $\EI$ function in Def. \ref{def:tssce} to exclude $\textit{ER}$ and solely rely  on $s(\alpha_{X \rightarrow Y})$. In the general case, when the entire time series is available, further explanations can also be generated in the future direction. For the retrospective case the definition remains unchanged. Further, we have to modify the valid Why Question, the Causal Scenario, and the Fundamental Rules, which were introduced for the Causal Hans example. It should also be mentioned that we  need to adjust the rules here partly because our game randomizes the sprites independently on the playing field. For this reason, individual rollouts vary in length regardless of the agent's behavior. In cases where rollouts are more comparable, population statistics can also be used.

\begin{defin}[Why Question - Behaviour] \label{def:dyntssceWhy}
Let $F_t$ be a frame of a rollout $R$, and let $x_t \in \text{Val}(X_t)$ be an instance of $X_t\in\mathbf{V}$ of the contextual-dependend SCM $\mathcal{M}$. We define $Q_{M,X_t} := x_t$ as a valid question if $X_t$ is a binary variable, $M$ is the appropriate contextual SCM for $F_t$, and $Q$ is true.
\end{defin}

\begin{defin}[Causal Scenario - Behaviour] \label{def:dyntssceCS}
The tuple $C_{XY}{:=}(\alpha_{X\rightarrow Y},x_t,y_k)$ is referred to as a behavioral causal scenario.
\end{defin}

\begin{defin}[Fundamental Rules - Behaviour] \label{def:dynamicTSSCEFundamentals}
Let $C_{XY}$ denote a causal scenario, let $s(x) \in \{-1,1\}$ be the sign of a scalar. We define FOL-based rule functions as 

\begin{enumerate}[start=1,label={(ER$\arabic*$})]
\item
$
\!
\begin{aligned}[t]
 \textnormal{If } x \in \{0,1\} \textnormal{, then: } 
    &(s(\alpha_{X {\rightarrow} Y}) < 0 \land x_t \oplus y_k) \vee  
    (s(\alpha_{X {\rightarrow} Y}) > 0 \land x_t == y_k)
\end{aligned}
$ 

\item
$
\!
\begin{aligned}[t]
\textnormal{If } x \in 	\mathbb{R}\textnormal{, then: } 
    &s(\alpha_{X {\rightarrow} Y}) 
\end{aligned}
$ 
\end{enumerate} indicating for each rule ER$i(\cdot)\in\{-1,0,1\}$ how the causal relation $X{\rightarrow} Y$ satisfies that rule.
\end{defin}

We previously described the three subprocesses for the Killer agent: (i) $C_{K,1} = \textnormal{`powerup and opponent exist'}$, (ii) $C_{K,2} = \textnormal{`powerup does not exist and opponent exists'}$, and (iii) $C_{K,3}=\textnormal{`neither exists'}$. In Fig. \ref{fig:basicCoinrunner}, for example, all sprites are present. The Killer agent would deterministically move towards the power-up since both the power-up and opponent exist. Afterward, a new subprocess directs the agent to move towards the opponent.

To obtain causal graphs for each context $C$, we first conditioned the rollouts from 500 runs of a nearly deterministic agent tailored to Killer behavior on the respective contexts. Here we also incorporated Margin Frames and a small amount of noise. For each of the 500 rollouts, we generated causal graphs using three methods: (i) Granger causality combined with Vector Autoregression (VAR), (ii) VARLiNGAM, and (iii) Lasso. The resulting Graphs for each method and context were individually averaged at the end.

For all methods, we considered a time lag of 1 and excluded instantaneous effects. First, we trained a VAR model for method (i) and conducted a Granger causality test for variables with p-values less than 0.05. We included the coefficients of the VAR model in the causal graphs when the significance level of the granger test was also less than 0.05. For method (ii), we ran the VARLiNGAM model without making significant changes to the default parameters. As for method (iii), we used a LassoCV implementation with 5-fold cross-validation.

\begin{figure}[h]
    \centering
    \begin{adjustbox}{max width=1.3\linewidth,center}
    \includegraphics[width=1.5\textwidth]{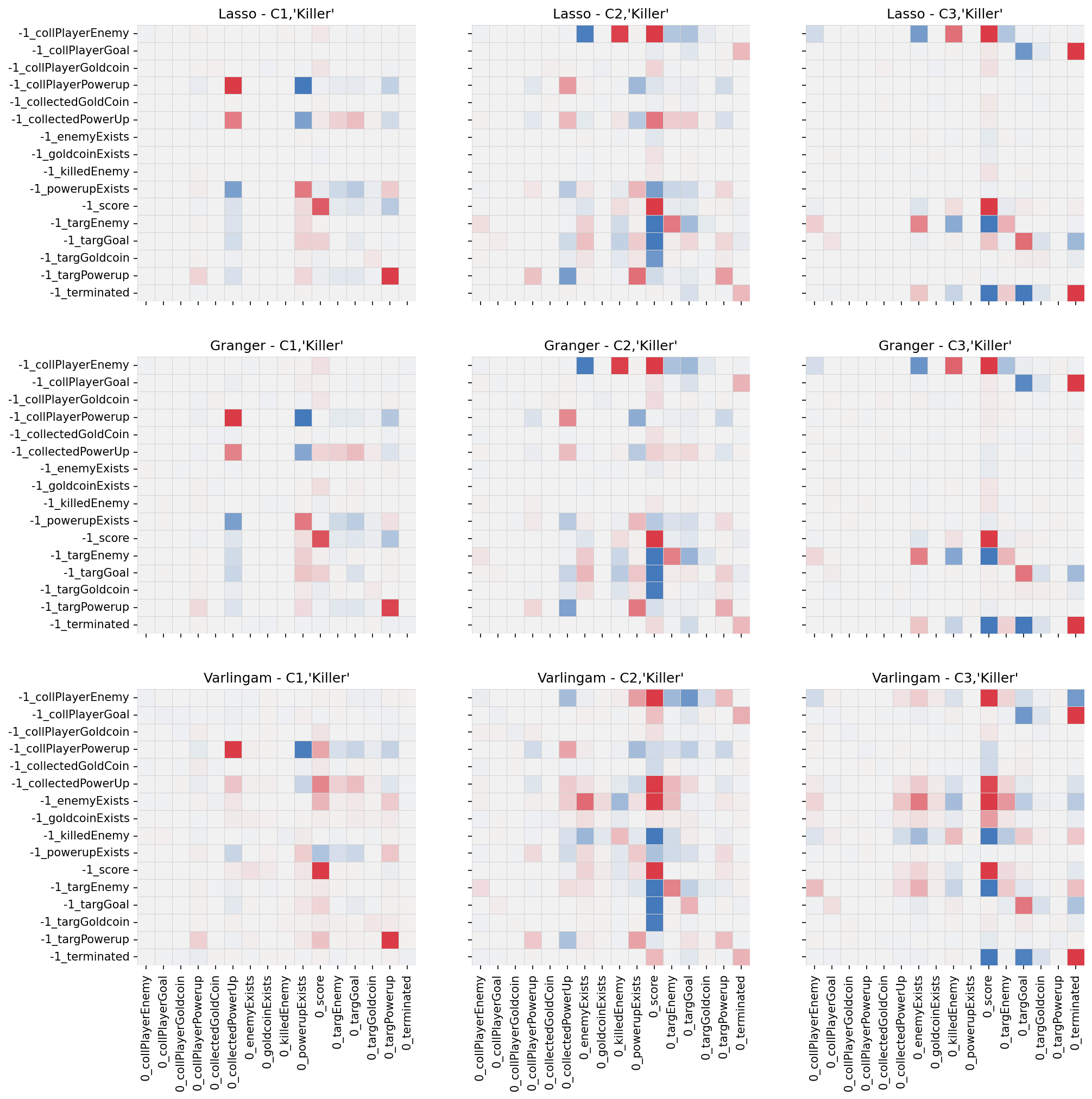}
    \end{adjustbox}
    \caption{\textbf{Comparative Causal Diagrams for Behaviour Type Killer Using Granger, Lasso, and VARLiNGAM for $C_{K,1}$, $C_{K,2}$, $C_{K,3}$.} Red indicates positive influences, while blue represents negative effects.}
    \label{fig:coinrunnerCausalGraphsOverview}
\end{figure}

Fig. \ref{fig:coinrunnerCausalGraphsOverview} displays the corresponding results. Rows with the prefix '-1\_' have a causal effect on columns with the prefix '0\_'. The variables used in the graphs are mostly self-explanatory and binary to provide meaningful explanations. 

It is important to note that the process of killing or picking up a sprite is divided into several steps in the time series. First, an object is ``targeted," then ``collided," and finally, it is no longer present or has been picked up in the next timestep. Additionally, we employed cosine similarity for target tracking, which offers an advantage over Manhattan Distance and Euclidean Norm, as it allows targeting only one sprite at a time.

Using the T-SCE approach, we can ask questions about why an agent performs a specific action in a given run if we consider the graphs to be causal. For example, in the main section, we answered the question, ``Why is Mario targeting the Goomba?". Fig. \ref{fig:CoinRunner_MainFigure} also shows what happens when Luigi, with the same behavior type, mistakenly targets the Goomba. Additionally, we compiled a collection of different answers from the three models to the questions: ``Why is Mario targeting the Goomba?" (Tab. \ref{tab:WhyDidMarioTargetTheGoomba}), ``Why did Mario jump on the Goomba?" (Tab. \ref{tab:WhyDidMarioJumpOnTheGoomba}), and ``Why did Mario run into the goal?" (Tab. \ref{tab:WhyDidMarioRunIntoTheGoal}) in the corresponding tables. We used a uniform expression of encoding to maintain comparability, and the response width for both the retrospective and anticipatory parts was uniformly limited. Anticipatory effects are limited to one time step ahead. 


\begin{table}[!h]
    \caption{Causal Explanation for the question`Why did Mario target the Goomba?' during a recorded rollout.}
    \label{tab:WhyDidMarioTargetTheGoomba}
\begin{tabularx}{\textwidth}{p{2cm}X}
  \toprule 
  \multicolumn{2}{c}{\textbf{Why did Mario target the Goomba?}}\\
  \midrule
  \textbf{Model} & \textbf{Explanation}\\
  \midrule 
  \multirow{3}{*}{Lasso} 
  & \textit{\textbf{Retrospective:}} Mario is targeting the enemy, constantly over action(s) 23 to 28, mostly because the enemy was targeted, because the player did not collide with the enemy, because the powerup did not exist and because the player had already collected the powerup continously in the previous time step. \vspace{1.5mm}\\
  & \textit{\textbf{Anticipative:}} Targeting the enemy now has a positive effect on targeting the enemy, the existence of the enemy, colliding with the enemy and a negative effect on the score, targeting the goal and killing the enemy in the next time step. \\
\midrule 
  \multirow{3}{*}{Varlingam} 
  &\textit{\textbf{Retrospective:}} Mario is targeting the enemy, constantly over action(s) 23 to 28, because the enemy was targeted, because the player did not collide with the enemy and because the enemy did exist continously in the previous time step. \vspace{1.5mm}\\ 
  & \textit{\textbf{Anticipative:}} Targeting the enemy has a positive effect on targeting the enemy, colliding with the enemy, collecting the powerup, the existence of the enemy and a negative effect on the score and targeting the goal in the next time step.\\
\midrule
  \multirow{3}{*}{GrangerVAR} 
  & \textit{\textbf{Retrospective:}} Mario is targeting the enemy, constantly over action(s) 23 to 28, mostly because the enemy was targeted, because the player did not collide with the enemy, because the player had already collected the powerup and because the powerup did not exist continously in the previous time step. \vspace{1.5mm}\\ 
  & \textit{\textbf{Anticipative:}} Targeting the enemy has a positive effect on targeting the enemy, the existence of the enemy and a negative effect on the score, targeting the goal, killing the enemy and targeting the goldcoin in the next time step.\\
  \bottomrule 
\end{tabularx}
\end{table}

\begin{table}[!h]
    \caption{Causal Explanation for the question`Why did Mario jump on the Goomba?' during a recorded rollout.}
    \label{tab:WhyDidMarioJumpOnTheGoomba}
\begin{tabularx}{\textwidth}{p{2cm}X}
  \toprule 
  \multicolumn{2}{c}{\textbf{Why did Mario jump on the Goomba?}}\\
  \midrule
  \textbf{Model} & \textbf{Explanation}\\
  \midrule 
  \multirow{3}{*}{Lasso} 
  & \textit{\textbf{Retrospective:}} Mario is colliding with the enemy because the enemy was targeted (0.056) due to the action before.  \vspace{1.5mm}\\
  & \textit{\textbf{Anticipative:}} Colliding with the enemy has a positive effect on the score (4.338), killing the enemy (0.483) and a negative effect on the existence of the enemy (-0.479), targeting the goal (-0.197), targeting the enemy (-0.182) and targeting the goldcoin (-0.029) in the next time step.\\
\midrule
  \multirow{3}{*}{Varlingam} 
  & \textit{\textbf{Retrospective:}} Mario is colliding with the enemy mostly because the enemy was targeted (0.059), because the player did not collide with the enemy (-0.02), because the enemy did exist (0.015) and altough the player did not collide with the goldcoin (0.014) through the action before. \vspace{1.5mm}\\
  & \textit{\textbf{Anticipative:}} Colliding with the enemy has a positive effect on the score (4.331), the existence of the powerup (0.23), targeting the powerup (0.15) and a negative effect on targeting the goal (-0.38), targeting the enemy (-0.235) and collecting the powerup (-0.219) in the next time step.\\
\midrule
  \multirow{3}{*}{GrangerVAR} 
  & \textit{\textbf{Retrospective:}} Mario is colliding with the enemy because the enemy was targeted (0.04) due to the action before. \vspace{1.5mm}\\
  & \textit{\textbf{Anticipative:}} Colliding with the enemy has a positive effect on the score (4.458), killing the enemy (0.486) and a negative effect on the existence of the enemy (-0.486), targeting the goal (-0.24), targeting the enemy (-0.201) and targeting the goldcoin (-0.04) in the next time step.\\
  \bottomrule 
\end{tabularx}
\end{table}

\begin{table}[t]
    \caption{Causal Explanation for the question`Why did Mario run into the Goal?' during a recorded rollout.}
    \label{tab:WhyDidMarioRunIntoTheGoal}
\begin{tabularx}{\textwidth}{p{2cm}X}
  \toprule 
  \multicolumn{2}{c}{\textbf{Why did Mario run into the Goal?}}\\
  \midrule
  \textbf{Model} & \textbf{Explanation}\\
  \midrule 
  \multirow{3}{*}{Lasso} 
  &\textit{\textbf{Retrospective:}} Mario is colliding with the goal because the goal was targeted (0.044) with the action before. \vspace{1.5mm}\\
  & \textit{\textbf{Anticipative:}} Colliding with the goal has a positive effect on termination of the game (0.518), the score (0.029) and a negative effect on targeting the goal (-0.377) and targeting the goldcoin (-0.041) in the next time step.\\
\midrule
  \multirow{3}{*}{Varlingam} 
  & \textit{\textbf{Retrospective:}} Mario is colliding with the goal mostly because the goal was targeted (0.055), because the player had killed the enemy (0.017), because the game is not terminated (-0.011) and altough the player had already collected the powerup (-0.015) the time step before. \vspace{1.5mm}\\
  & \textit{\textbf{Anticipative:}} Colliding with the goal now has a positive effect on termination of the game (0.505), killing the enemy (0.012) and a negative effect on targeting the goal (-0.374), targeting the goldcoin (-0.055), the existence of the enemy (-0.018) and targeting the enemy (-0.014) in the next time step.\\
  \midrule
  \multirow{3}{*}{GrangerVAR} 
  &\textit{\textbf{Retrospective:}} Mario is colliding with the goal mostly because the goal was targeted (0.023) and because the goldcoin did exist (0.01) the time step before. \vspace{1.5mm}\\ 
  & \textit{\textbf{Anticipative:}} Colliding with the goal now has a positive effect on termination of the game (0.577), the score (0.026) and a negative effect on targeting the goal (-0.434) and targeting the goldcoin (-0.054) in the next time step.\\
  \bottomrule 
\end{tabularx}
\end{table}

\clearpage
\section{Towards Contexting T-SCE with Existing Causal XAI Paradigms}
We compared our method with two other markedly distinct methods. Additional information pertaining to the derivation of the results is consolidated and presented herein.

The authors of the Causal Shapley Values have published an R implementation, which we used for our test. Based on the classification dataset, we trained an XGBoost model with about 500,000 samples for the Causal Shapley Values. Of these, 85\% were used for training, 10\% for validation, and the remaining samples for creating graphs and as a basis for distinguishing between the predicted groups `below average' and `above average' mobility. An XGBoost classifier was trained using 5-fold cross-validation (early stopping at 50, 5000 rounds, maximum depth 10, eta 0.3).

\begin{figure}[t]
    \centering
    \includegraphics[width=1\textwidth]{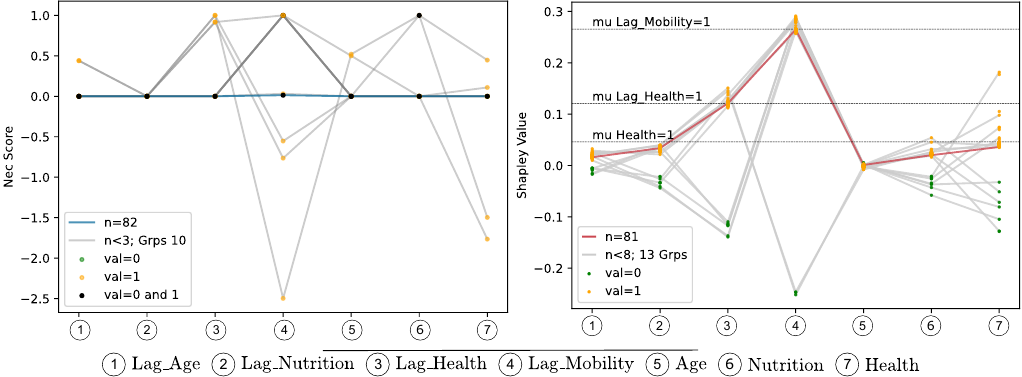}
    \caption{Necessity Scores for all features of 100 patients from the `above average' Mobility group. } \label{fig:contextjointappendix}
\end{figure}

In addition to the symmetrical Causal Shapley Values for the 'below average' group, we also produced a plot for the 'above average' group. Fig. \ref{fig:contextjointappendix} displays the nearly perfectly symmetrical Causal Shapley Values for the 'above average' group.

The authors of the LEWIS method were kind enough to provide us with a part of their code for the causality-based counterfactual method. Similar to the Shapley method, a classification model was initially trained. This time a Random Forest approach was used, with 15 estimators and a maximum depth of five, and max\_features set to `sqrt'. For the `above average' group, we have compiled the corresponding importance of the patients' features also in Fig. \ref{fig:contextjointappendix}. In this case, these are the Necessity Scores from local explanations, where the change from the current category to the alternative category was considered.

\section{Towards XIL and Causal XIL}
\begin{figure*}[h]
    \centering
    \includegraphics[width=1\textwidth]{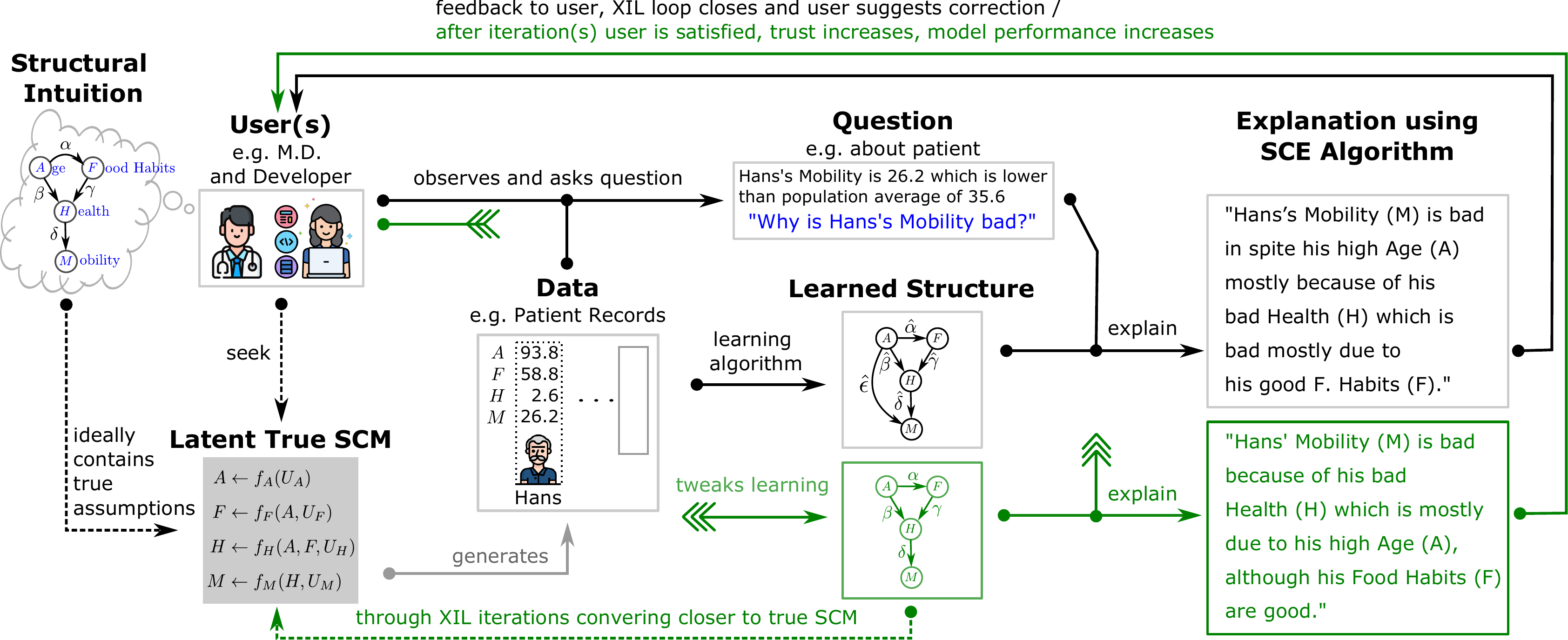}
    \caption{\textbf{Causal XIL Paradigm in action.} A practical example featuring a developer and a doctor collaboratively working on optimizing a causal model. The feedback loop enables iterative improvement of the model, drawing on the expert knowledge of the doctor and the adjustments made by the developer \citep{anonymous2023causal}.} \label{fig:causalXILSCE}
\end{figure*}

According to the article on the Explanatory Interactive Machine Learning (XIL) paradigm \citep{XILOriginal}, it aims to bridge the gap between interactive learning and explainability in order to foster trust in machine learning models. This is achieved by focusing on the interactions between the learner and the user. During these interactions, the user is able to verify, correct, and provide feedback on the model's predictions and explanations. This direct involvement of the user enables a better understanding of the model's decisions and promotes more effective learning by adapting the learning process to the needs and expectations of the user. The article points out that existing approaches, such as passive learning or interactive learning, usually do not address the aspect of 'trust,' and do not employ explanation algorithms, resulting in the model being a black box for the user. In these cases, the user cannot directly verify why a prediction was made based on a query. Consequently, the user can never be sure whether correct predictions were made, or if they were merely coincidental.

The proposed approach is as follows: The learner selects a sample and classifies it. Along with a local explanation (i.e., LIME), the sample is presented to the user. The user can see both and validate whether the correct decision was made for the right reasons or not. At the same time, the user can provide feedback on erroneous behavior to the learner. This would be the case, for example, if the learner makes the correct decision for the wrong reasons or makes the wrong decision for the right reasons. If the prediction is incorrect for the wrong reasons, the user provides the correct label to the learner. The case of being correct for the wrong reasons is more interesting; here, the user must improve the explanation and provide it back to the model. As a result, the user can specifically guide the model and have a direct influence on its behavior. CAIPI, a model-independent instance of the paradigm, was introduced by the authors in the paper.

\cite{anonymous2023causal} take their work a step further in their article, describing a causal variant of the XIL paradigm. They argue that the need for a causal variant lies in avoiding deceptive correlations and fallacies, also referred to as ``spurious fallacies," in explanation methods that are not truly causal. A causal approach enables more effective collaboration between humans and AI by providing explanations that better align with human logic and causality. Since humans often establish causal relationships in their thinking, a causal variant of the XIL paradigm allows for a better translation between human thought and AI models. Anchoring the model's explanations in a causal model, combined with a feedback loop, would both increase trust through understanding and transparency, as well as improve the robustness of the models by uncovering and correcting weaknesses simultaneously.

Fig. \ref{fig:causalXILSCE} shows a structural diagram of the causal XIL paradigm using a practical example. On the left side, two actors are depicted in the diagram: a developer and a doctor (expert). Both individuals have access to a dataset of patient records in this example. The developer can generate an initial causal model from the data using a Causal Discovery method. Since the doctor already has a mental representation of the causal model underlying the dataset, he serves as a verifier in this case.

Assume that the doctor examines a dataset for the patient, Hans, who has lower mobility than average. The doctor asks, ``Why is Hans' mobility so poor?". Based on the already learned causal model, a causal answer to this question can be generated. According to the XIL paradigm, the explanation can be either correct or incorrect. If the explanation is incorrect, the doctor informs the developer, who then retrains the model considering the modified dataset.

The doctor can now ask the question again and receive an improved answer through the feedback loop. The learned causal model is now closer to the true latent Structural Causal Model (SCM). It is essential to note that although the answer in this example with four variables seems manageable, a direct translation of the doctor's mental model may not be straightforward with a larger number of variables.

\cite{anonymous2023causal} not only introduced the causal variant of the XIL paradigm but also proposed an approach to generate causal explanations from a causal diagram for initial application areas, the Structural Causal Explanation algorithm.

\end{document}